\documentclass[conference]{IEEEtran}
\IEEEoverridecommandlockouts
\usepackage{cite}
\usepackage{amsmath,amssymb,amsfonts}
\usepackage{algorithmic}
\usepackage{graphicx}
\usepackage{textcomp}
\usepackage{xcolor}
\usepackage{hyperref}
\usepackage{titlesec}
\usepackage{booktabs}
\usepackage{array} 
\usepackage{float}

\usepackage{subcaption}
\titleclass{\subsubsubsection}{straight}[
\subsubsection]
\newcounter{subsubsubsection}[subsubsection]
\renewcommand\thesubsubsubsection{\thesubsubsection.\arabic{subsubsubsection}}
\titleformat{\subsubsubsection}[runin]
  {\normalfont\normalsize\bfseries}{\thesubsubsubsection}{1em}{}
\titlelabel{\thetitle.\quad}
\def\BibTeX{{\rm B\kern-.05em{\sc i\kern-.025em b}\kern-.08em
    T\kern-.1667em\lower.7ex\hbox{E}\kern-.125emX}}
\begin{document}

\title{MobileEgo Anywhere: Open Infrastructure for long horizon egocentric data on commodity hardware\\
}

%\author{\IEEEauthorblockN{Anonymized Author*}
%\IEEEauthorblockA{\textit{Anonymized Company} \\
%XXXX, YYYY \\
%email-id}}

\author{
    \IEEEauthorblockN{
        Senthil Palanisamy*, 
        Abhishek Anand*, 
        Satpal Singh Rathore*, \\
        Pratyush Patnaik*, 
        Shubhanshu Khatana*, 
        Ekaksh Janweja*
    }
    \IEEEauthorblockA{
        \textit{FPV Labs} \\
        Emails: abhishek@fpvlabs.ai
    }

\thanks{*All authors contributed equally to this work.}
}
\maketitle

\begin{abstract}
Vision-language-action (VLA) models have driven demand for large-scale egocentric datasets, yet the hardware and infrastructure to collect long-horizon data remain inaccessible. Datasets today typically have episodes only a few minutes long, which fails to capture the long-horizon temporal dependencies that complex robotic task execution requires. We present \textbf{MobileEgo Anywhere}, a framework for collecting hour-plus egocentric trajectories on commodity mobile hardware that uses modern smartphone sensors for long-term pose tracking without the hardware barriers of traditional robotics data collection. We release three components: (1) STERA, an open-source video-processing pipeline that converts raw mobile captures into standardized, training-ready formats for VLA and foundation-model research; (2) a free mobile app that lets any user record egocentric activity; and (3) a 200-hour dataset of diverse, long-form egocentric data with persistent state tracking across 584 sessions. We further show this data is a usable training signal: mid-training a VLA on it lowers held-out action-prediction error.
\begin{IEEEkeywords} VLA training dataset, egocentric dataset, robotics, commoditized VLA data collection, long-horizon tracking. \end{IEEEkeywords}
\end{abstract}

%===============================================================================

\begin{figure*}[t] % Use figure* for two-column spanning. Note: [H] does not work well with figure*
    \centering
    \begin{subfigure}[b]{0.22\textwidth} % Slightly adjusted widths to fit the wider page nicely
        \centering
        \includegraphics[width=\textwidth]{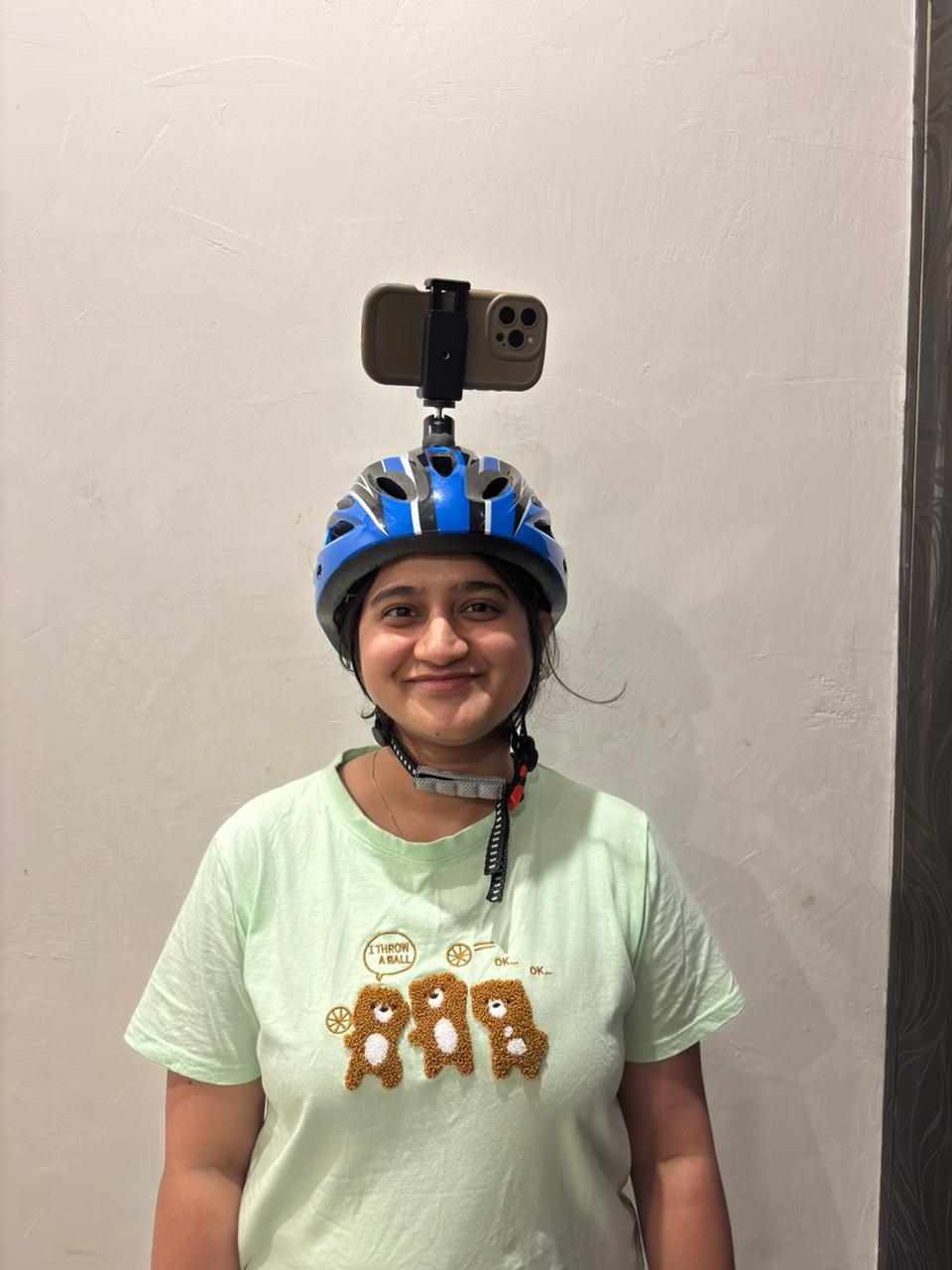}
        \caption{MobileEgo Anywhere recording setup.}
        \label{fig:hero_left}
    \end{subfigure}
    \hfill
    \begin{subfigure}[b]{0.42\textwidth}
        \centering
        \includegraphics[width=\textwidth]{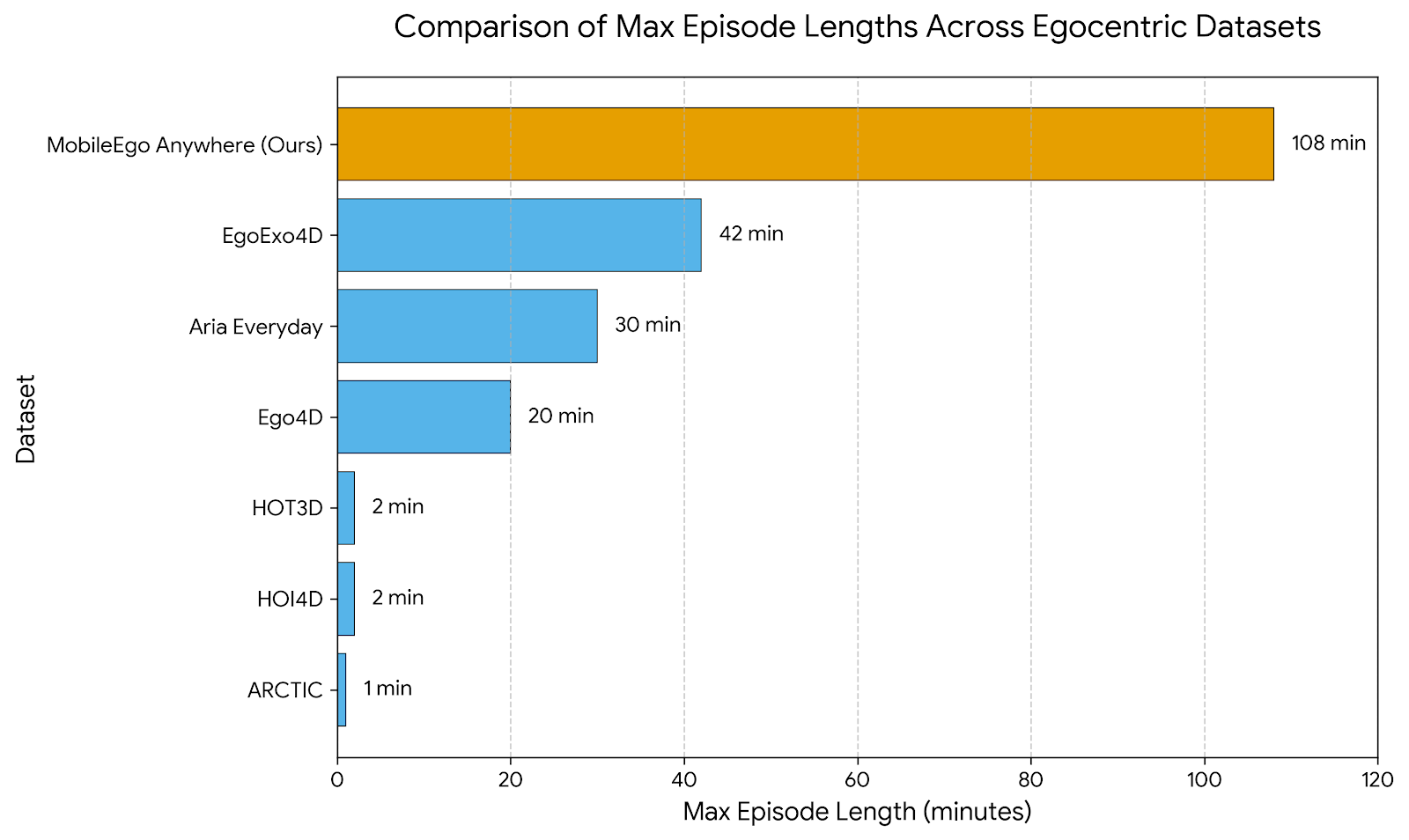}
        \caption{Comparison of episode duration.}
        \label{fig:hero_center}
    \end{subfigure}
    \hfill
    \begin{subfigure}[b]{0.32\textwidth}
        \centering
        \includegraphics[width=\textwidth]{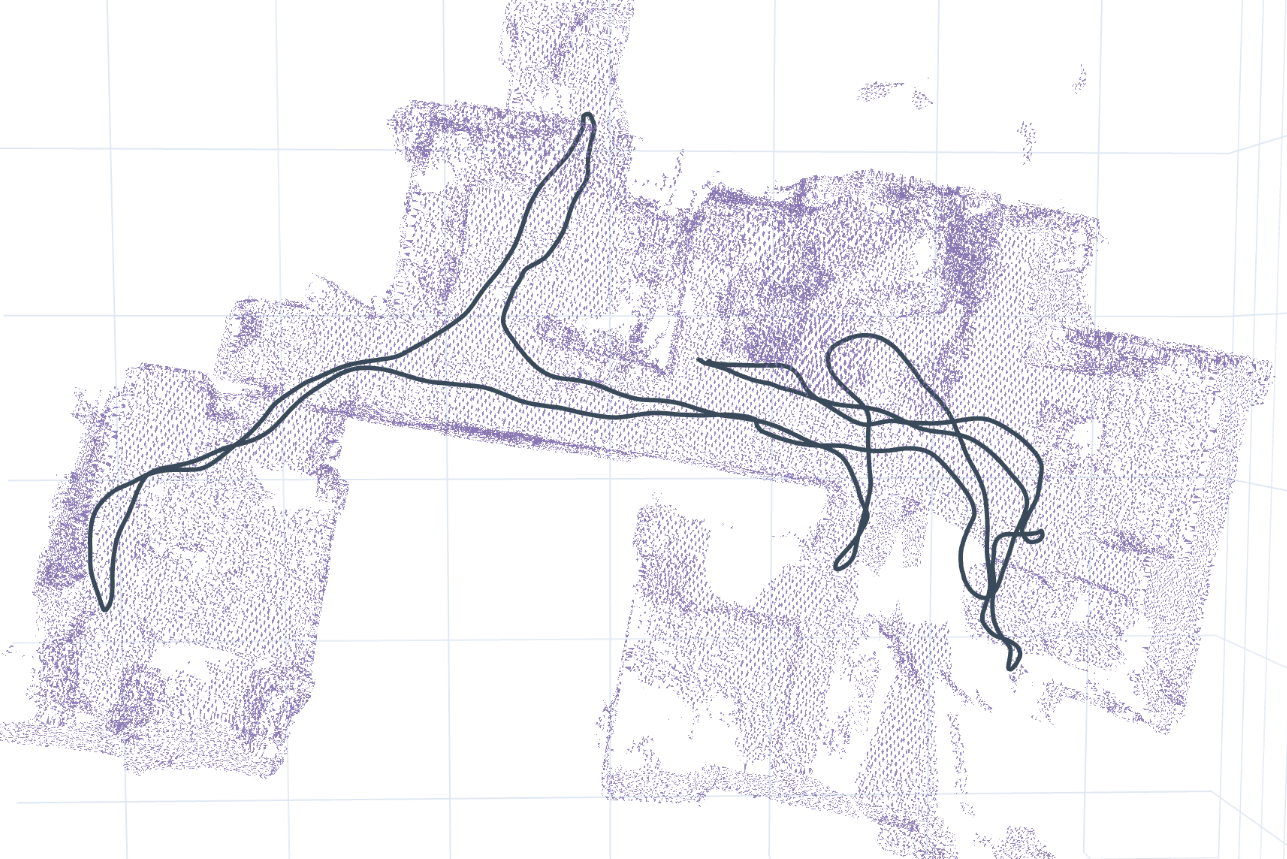}
        \caption{Long-horizon trajectory tracked from ARKit.}
        \label{fig:hero_right}
    \end{subfigure}
    \caption{\textbf{MobileEgo Anywhere} turns any modern iPhone into a long-horizon egocentric capture device. (a)~Contributors record hands-free using a helmet-mounted phone. (b)~Episodes are substantially longer than those in prior datasets. (c)~ARKit-based visual-inertial fusion yields continuous 6\,DoF pose, used to generate 3D hand trajectories in a consistent world frame across the full session.}
    \label{fig:hero_main}
\end{figure*}

\section{Introduction}

Vision-language-action (VLA) models have rapidly become a leading approach to general-purpose robot policies. Zheng et al.~\cite{zheng2026egoscale} establish a log-linear scaling law $L = 0.024 - 0.003 \times \ln(D)$ between the volume of egocentric human data $D$ (in hours) and validation loss $L$, implying that further gains require roughly an order-of-magnitude more data across more environments.

VLA training draws on a hierarchy of data sources. Internet video offers broad semantic coverage but no contact dynamics. Simulation scales cheaply, yet policies trained in it still face a sim-to-real gap. Egocentric human video and interfaces such as UMI~\cite{chi2024umi} provide richer interaction signal and are a primary source for large-scale pretraining, which needs a large, varied corpus spanning many environments and long-horizon tasks.

Existing egocentric datasets share two limits: short episodes and high collection-hardware barriers. MobileEgo Anywhere addresses both by using the visual-inertial odometry (VIO) already built into modern smartphones, specifically ARKit on the iPhone Pro, for 6\,DoF pose tracking with no specialized peripherals. We release a free capture app, an open-source processing suite (STERA), and a 200-hour dataset of household activities with continuous episodes up to 108 minutes, 3D hand trajectories, and three-level hierarchical language annotations. We validate the pipeline along three axes: ARKit pose accuracy against motion-capture ground truth, ground-truth-free hand-pose consistency, and hierarchical label quality, then show the data serves as a usable training signal for a vision-language-action model.

\section{Related Work}

\subsection{Egocentric Datasets for Robotics}
Early egocentric datasets primarily focused on action recognition and localized human-object interactions. Large-scale efforts such as Ego4D~\cite{grauman2022ego4d} and EPIC-KITCHENS~\cite{damen2018scaling,damen2022rescaling} provided the community with thousands of hours of video, but these were largely passive and lacked the precise, continuous 6\,DoF pose tracking required for robotic policy learning. Recent shifts toward  Foundation Models and VLA architectures \cite{egovlp} \cite{egomimic} have increased the demand for actionable egocentric data. Projects like EgoScale~\cite{zheng2026egoscale} provide precise poses but episodes are very short. EgoExo4D~\cite{grauman2024egoexo4d} adds pose, depth, and hand annotations but relies on Meta's Project Aria glasses and synchronized exo cameras, hardware that is not commercially available. HOI4D~\cite{liu2022hoi4d}, ARCTIC~\cite{fan2023arctic}, and HOT3D~\cite{banerjee2024hot3d} provide precise hand tracking but cover only seconds to minutes in controlled lab settings. EgoDex~\cite{egodex} offers 829 hours of dexterous manipulation data captured on Apple Vision Pro, but with $\sim$9\,s demonstrations. We instead collect long-horizon trajectories that stay spatially consistent across hour-plus sessions, using only consumer hardware.

\subsection{Scalable Data Collection Interfaces}
Teleoperation and kinesthetic teaching yield high-quality samples but scale poorly, since each demonstration needs an operator and a robot. Open X-Embodiment~\cite{open_x} and DROID~\cite{DROID} assemble large teleoperation datasets. The Universal Manipulation Interface (UMI)~\cite{chi2024umi} utilizes handheld grippers to bridge the gap between human demonstration and robotic execution, lowering the hardware barrier, but still requires specialized physical mounts and calibrated setups. Our approach instead leverages the commodity smartphone as a universal sensor suite, enabling collection anywhere without additional peripherals.

\subsection{Long-Term Egocentric SLAM and State Estimation}
Maintaining stable state tracking over long sessions is the central difficulty for SLAM in this setting. Structure-from-Motion pipelines such as COLMAP~\cite{colmapsfm} become computationally intractable on hour-long trajectories, while feature-based SLAM such as ORB-SLAM3~\cite{ORBSLAM3} accumulates drift in dynamic or texture-poor indoor scenes. Recent mobile AR frameworks, notably ARKit and ARCore, address this by integrating high-frequency IMU data with visual keyframes, enabling robust long-term tracking on edge devices.

\section{System Overview}

\begin{figure}[t]
    \centering
    \includegraphics[width=0.5\textwidth]{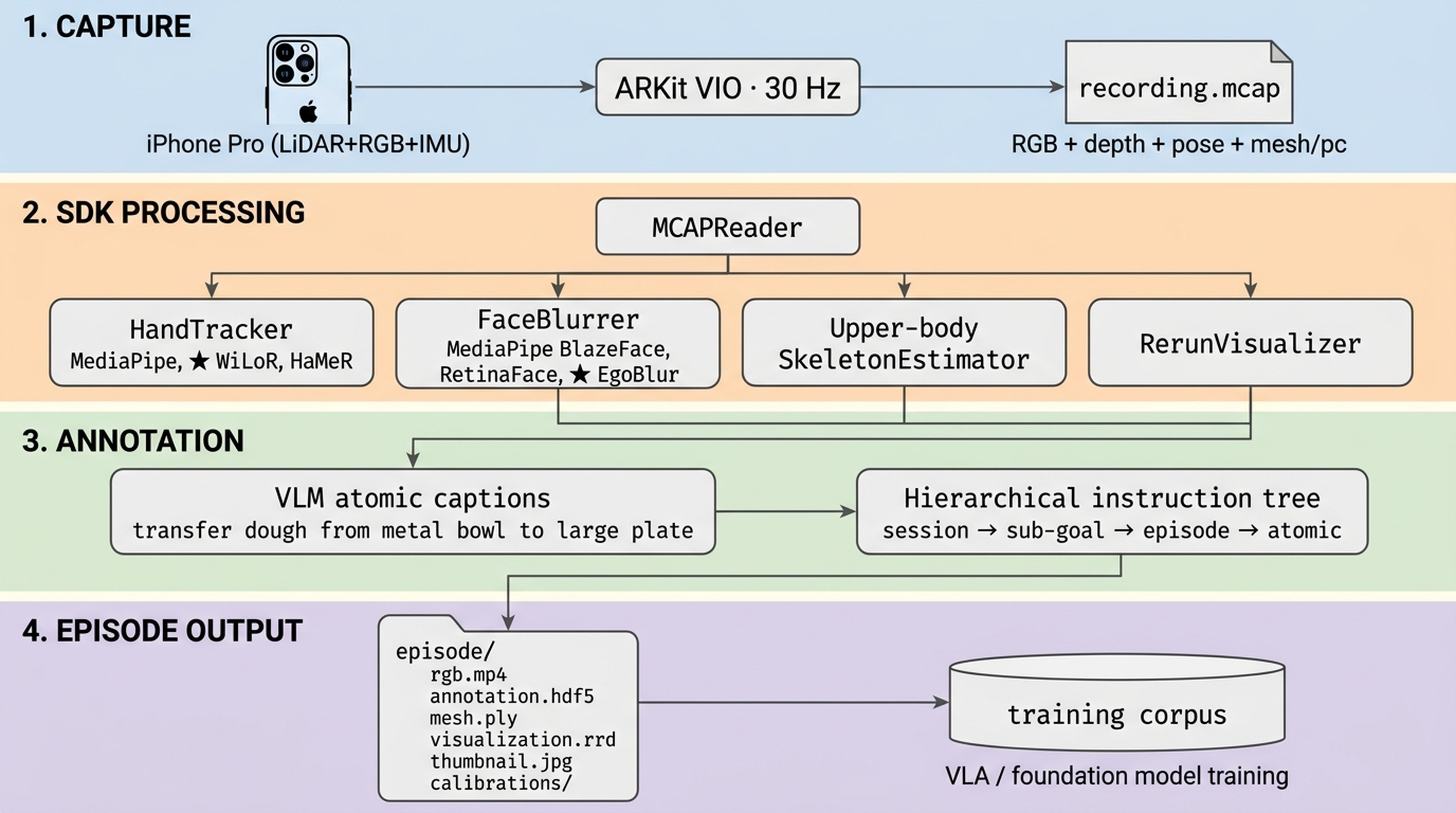}
    \caption{Overall data flow: raw mobile capture (RGB-D, IMU, ARKit pose) is logged in MCAP format, then processed offline into 3D hand trajectories, atomic action labels, and a hierarchical instruction tree.}
    \label{fig:process}
\end{figure}

Our hardware configuration uses a LiDAR-enabled iPhone Pro mounted on a head-worn rig for a first-person perspective of the participant's hands and workspace. ARKit captures synchronized RGBD streams providing 6\,DoF camera poses and per-frame depth maps; a dedicated mobile app exports raw sensor data (RGBD frames, IMU readings, camera intrinsics) in MCAP format~\cite{MCAP}. The offline STERA pipeline then transforms these logs into 3D hand trajectories, atomic action labels, and hierarchical task instructions.\footnote{Project resources: (1) Mobile App: \url{https://apps.apple.com/in/app/stera-by-fpv-labs/id6756263398}; (2) Python Processing Suite: \url{https://github.com/fpv-labs/stera-sdk}; (3) Huggingface Dataset: \url{https://huggingface.co/datasets/fpvlabs/stera-10m}; (4) Data Visualization: \url{https://platform.fpvlabs.ai/dataset/stera-10m/viz}};
\subsection{Capture Methodology}

Contributors secure the iPhone to a head-worn mount positioned for a consistent egocentric field of view; while a standard helmet mount was used here, any mounting hardware providing sufficient elevation is compatible. Data collection is managed via an integrated voice command interface (``start''/``stop'' triggers), ensuring hands-free operation critical for naturalistic activities. During recording, ARKit performs real-time sensor fusion, generating 6\,DoF camera poses by synchronizing the onboard IMU with the RGBD stream. The application archives RGB frames, depth maps, and IMU metadata registered to a common high-resolution timestamp, ensuring temporal consistency across all modalities for downstream 3D reconstruction and action recognition.

\subsection{3D Hand Trajectory Estimation}
\label{3d_hand}

High-fidelity 3D hand trajectories map human motion to robot end-effector frames via IK for VLA training. We employ WiLoR~\cite{potamias2024wilor} with MANO parameterization~\cite{romero2017mano} to estimate hand joints under biomechanical constraints, handling the partial occlusions common in first-person manipulation more reliably than alternatives such as MediaPipe~\cite{mediapipe}. WiLoR's relative 3D coordinates are localized into a global frame by sampling ARKit depth maps at detected joint locations and applying the extrinsic camera transformation, yielding world-anchored trajectories for imitation learning.

\subsection{Atomic Action Labels}

Action-conditioned VLA policies require language labels specifying \textit{which} object is manipulated, \textit{what} the action is, and \textit{where} the object moves, details that generic labels like ``pick up object'' do not provide. To produce labels at this level of specificity across 200 hours of video, we employ an automated annotation pipeline. Raw video is partitioned into contiguous, non-overlapping temporal spans, and each span is processed by a VLM that outputs a short imperative sentence constrained to include object modifiers (color, material, size) and spatial prepositions wherever the video evidence supports them (e.g., ``transfer dough from metal bowl to large plate'').

Validation against 50 human-annotated sessions shows automated labels average 7.95 words versus 2.94 and 1.09 descriptive modifiers per label versus 0.09. The automated pipeline produced zero temporal defects across all 5,249 labels. Human annotations contained 63 segments with durations $\leq$0\,s and 877 overlapping consecutive pairs (9.9\% of 8,821 adjacent pairs), defects that would propagate as corrupted training samples.

\subsection{Hierarchical Task Instructions}
\label{subsubsection:Hierarchical Task Instructions}

Long-horizon sessions contain dozens of atomic labels belonging to distinct sub-tasks. To expose this structure, atomic span captions are organized into a three-level instruction tree: a session-level goal, sub-goals, and episodes. Hierarchical instruction setups like these have been well studied in works like \cite{alfred} and \cite{canSay}. A language model groups temporally contiguous spans into episodes, clusters related episodes into sub-goals, and synthesizes a session-level goal grounded in the concrete objects across all spans.

Three invariants are enforced: unique span assignment, exact timestamp boundaries, and full session coverage with no gaps. Of seven language models evaluated, six produced fully valid outputs, providing language conditioning from 5-second manipulation steps to full session plans, matching the multi-scale supervision used by recent hierarchical VLA architectures.

\section{Data Quality Validation}

\begin{table*}[!t]
\centering
\caption{Comparison of egocentric datasets for robot-relevant pretraining. Hours and durations are taken from original publications or computed directly from reported totals.}
\label{tab:comparison}
\small
\setlength{\tabcolsep}{4pt}
\begin{tabular*}{\textwidth}{@{\extracolsep{\fill}}lcccccl@{}}
\toprule
\textbf{Dataset} & \textbf{Hours} & \textbf{Max Episode} & \textbf{6\,DoF Pose} & \textbf{Depth} & \textbf{Hand Annot.} & \textbf{Capture Hardware} \\
\midrule
Ego4D~\cite{grauman2022ego4d}               & 3{,}670     & up to $\sim$7\,hrs       & Partial & No  & Partial & Mixed (GoPro, ZShades, Aria) \\
EPIC-KITCHENS-100~\cite{damen2022rescaling} & 100         & $\sim$8.6\,min avg       & No      & No  & Partial & Head-mounted GoPro \\
EgoExo4D~\cite{grauman2024egoexo4d}         & 1{,}286     & $\sim$42\,min            & Yes     & Yes & Yes     & Aria + exo cameras \\
HOI4D~\cite{liu2022hoi4d}                   & $\sim$22.2  & $\sim$20\,sec / clip     & Yes     & Yes & Yes     & Intel RealSense \\
HOT3D~\cite{banerjee2024hot3d}              & $\sim$13.9  & $\sim$2\,min / recording & Yes     & No  & Yes     & Aria + Quest~3 \\
ARCTIC~\cite{fan2023arctic}                 & $\sim$2.1   & $\sim$23\,sec avg        & Yes     & No  & Yes     & MoCap rig \\
Aria Everyday~\cite{lv2024aria}             & $\sim$7.3   & $\sim$3\,min avg         & Yes     & No  & Yes     & Project Aria \\
EgoDex~\cite{egodex}                        & 829         & $\sim$9\,sec / demo      & Yes     & No  & Yes     & Apple Vision Pro \\
\midrule
\textbf{MobileEgo Anywhere (ours)}          & \textbf{200} & \textbf{108\,min}       & \textbf{Yes} & \textbf{Yes} & \textbf{Yes} & \textbf{Consumer iPhone} \\
\bottomrule
\end{tabular*}
\end{table*}

The released dataset contains 584 sessions totaling 200 hours from 20 contributors, averaging 20.5 minutes with a maximum of 108 minutes. Table~\ref{tab:comparison} shows MobileEgo Anywhere is the only dataset combining consumer hardware with continuous 6\,DoF pose, LiDAR depth, MANO annotations, and sessions exceeding one hour; EgoExo4D offers similar modalities but requires commercially unavailable hardware.

\subsection{ARKit Pose Accuracy}

\subsubsection{Motion Capture Ground Truth Comparison}
\label{mocap_arkit}
To quantify the absolute accuracy of ARKit pose estimates, we collected trajectories spanning representative motion profiles and evaluated them against ground truth from a 30-camera Vicon motion capture system. Tracked trajectories include two naturalistic egocentric sequences in which a participant performs everyday household tasks, a slow walk, a closed-loop traversal, and a spinning sequence. Absolute Trajectory Error (ATE RMSE), relative ATE, and Relative Pose Error (RPE)~\cite{sturm2012benchmark} for translation are reported in Table~\ref{tab:trajectory_data_updated}.

Table~\ref{tab:trajectory_data_updated} reports per-sequence ATE RMSE, relative ATE, and translational and rotational RPE against the Vicon ground truth. Relative ATE stays below 1\% for nine of the ten sequences and rotational RPE below 4$^\circ$; the lone exception is the short spinning sequence, whose elevated values follow from its rapid rotational motion. Translational RPE remains below 5\,cm throughout, so local pose consistency holds even where global drift is marginally elevated. ARKit visual-inertial odometry on a consumer iPhone Pro thus provides trajectory accuracy sufficient for the world-frame hand-pose anchoring of Section~\ref{3d_hand}.

\begin{table}[!t]
\centering
\caption{ARKit accuracy vs.\ Vicon ground truth}
\label{tab:trajectory_data_updated}
\resizebox{\columnwidth}{!}{%
\begin{tabular}{clccccc}
\hline
\textbf{Sequence No.} & \textbf{Duration} & \textbf{Traj.} & \textbf{ATE RMSE} & \textbf{Rel.\ ATE} & \textbf{RPE trans} & \textbf{RPE rot} \\ \hline
1  & 149.7\,s & 107.3\,m & 15.0\,cm & 0.14\% & 4.99\,cm & 3.4033$^\circ$ \\
2  & 134.5\,s &  41.6\,m &  9.1\,cm & 0.22\% & 2.24\,cm & 0.8529$^\circ$ \\
3  & 134.4\,s &  51.6\,m & 10.1\,cm & 0.20\% & 2.59\,cm & 0.5103$^\circ$ \\
4  &  47.1\,s &   6.9\,m & 10.1\,cm & 1.47\% & 2.04\,cm & 4.1728$^\circ$ \\
5  &  91.5\,s &  28.1\,m & 13.0\,cm & 0.46\% & 4.03\,cm & 3.6004$^\circ$ \\
6  & 120.7\,s &   0.03\,m &  0.01\,cm & 0.48\% & 0.01\,cm & 0.0468$^\circ$ \\
7  & 123.4\,s &   0.04\,m &  0.04\,cm & 0.98\% & 0.01\,cm & 0.0644$^\circ$ \\
8  & 163.7\,s &  40.1\,m & 10.1\,cm & 0.25\% & 3.27\,cm & 1.6832$^\circ$ \\
9  & 145.8\,s &  37.2\,m & 11.9\,cm & 0.32\% & 2.75\,cm & 1.6550$^\circ$ \\
10 &  56.1\,s &  32.5\,m & 12.6\,cm & 0.39\% & 1.76\,cm & 1.6650$^\circ$ \\
\hline
\end{tabular}%
}
\end{table}

\subsubsection{Long-Term Drift Evaluation}
Because ARKit is closed-source, we assess long-term drift by placing an ArUco marker~\cite{garridojurado2014aruco} at session start and revisiting it at roughly the temporal midpoint and session end. We repeat this across six environments (Table~\ref{tab:sighting_experiments}); drift is below 1\,cm in all but the whole-house traversal (1.5\,cm end-of-session) and below 0.1\% of trajectory length in all cases, demonstrating the efficacy of ARKit tracking for downstream VLA applications.

\begin{table}[!t]
    \centering
    \caption{ARKit long-term drift evaluation via ArUco marker revisits.}
    \label{tab:sighting_experiments}
    \begin{tabular}{lcc}
        \toprule
        \textbf{Environment} & \textbf{Mid-session} & \textbf{End-of-session} \\
        \midrule
        Kitchen activity         & 0.4\,cm & 0.7\,cm \\
        Living-space activity    & 0.3\,cm & 0.4\,cm \\
        Whole-house activity     & 1.0\,cm & 1.5\,cm \\
        Whole-house walk-through & 0.3\,cm & 0.2\,cm \\
        Cloth folding            & 0.1\,cm & 0.1\,cm \\
        Nail polish              & 0.1\,cm & 0.1\,cm \\
        \bottomrule
    \end{tabular}
\end{table}

%% Tree-example figure (fig:tree_example) relocated to the appendix to save main-body space.

\section{Training Signal Validation}

\begin{figure*}[t]
    \centering
    \begin{subfigure}[t]{0.64\textwidth}
        \centering
        \includegraphics[width=\textwidth]{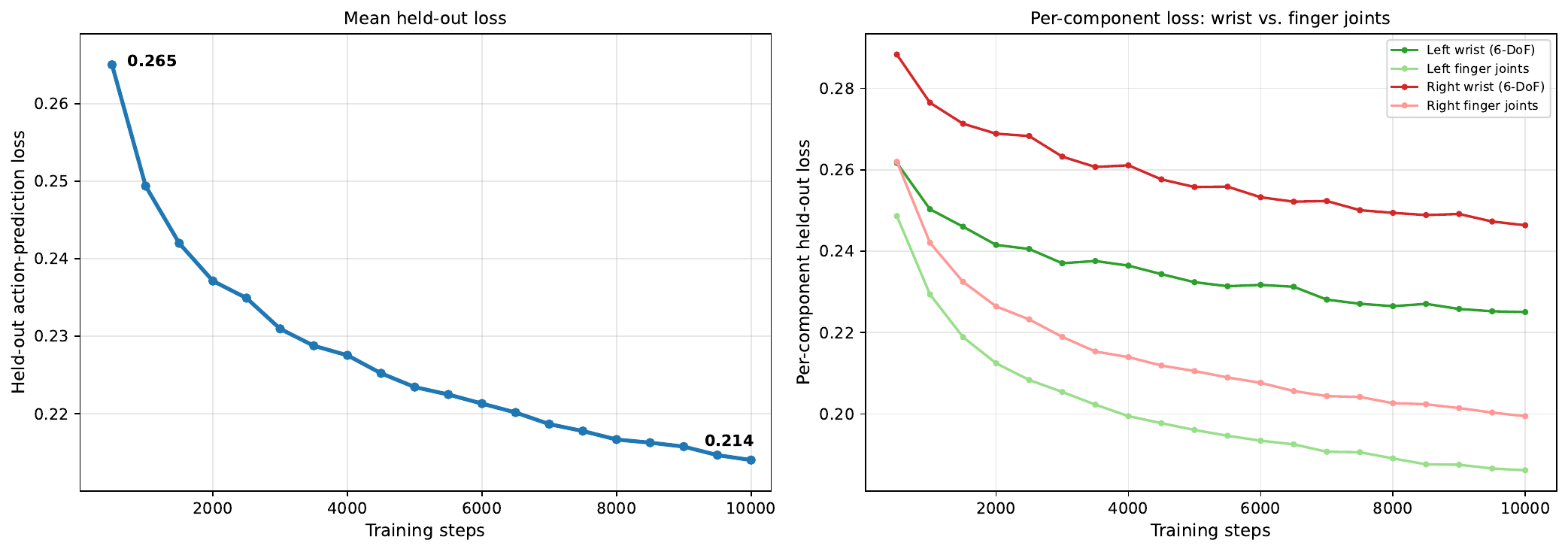}
        \caption{Held-out action-prediction loss over 10{,}000 mid-training steps ($-19\%$); the reduction is driven by finger articulation rather than the wrist root.}
        \label{fig:val_training}
    \end{subfigure}
    \hfill
    \begin{subfigure}[t]{0.34\textwidth}
        \centering
        \includegraphics[width=\textwidth]{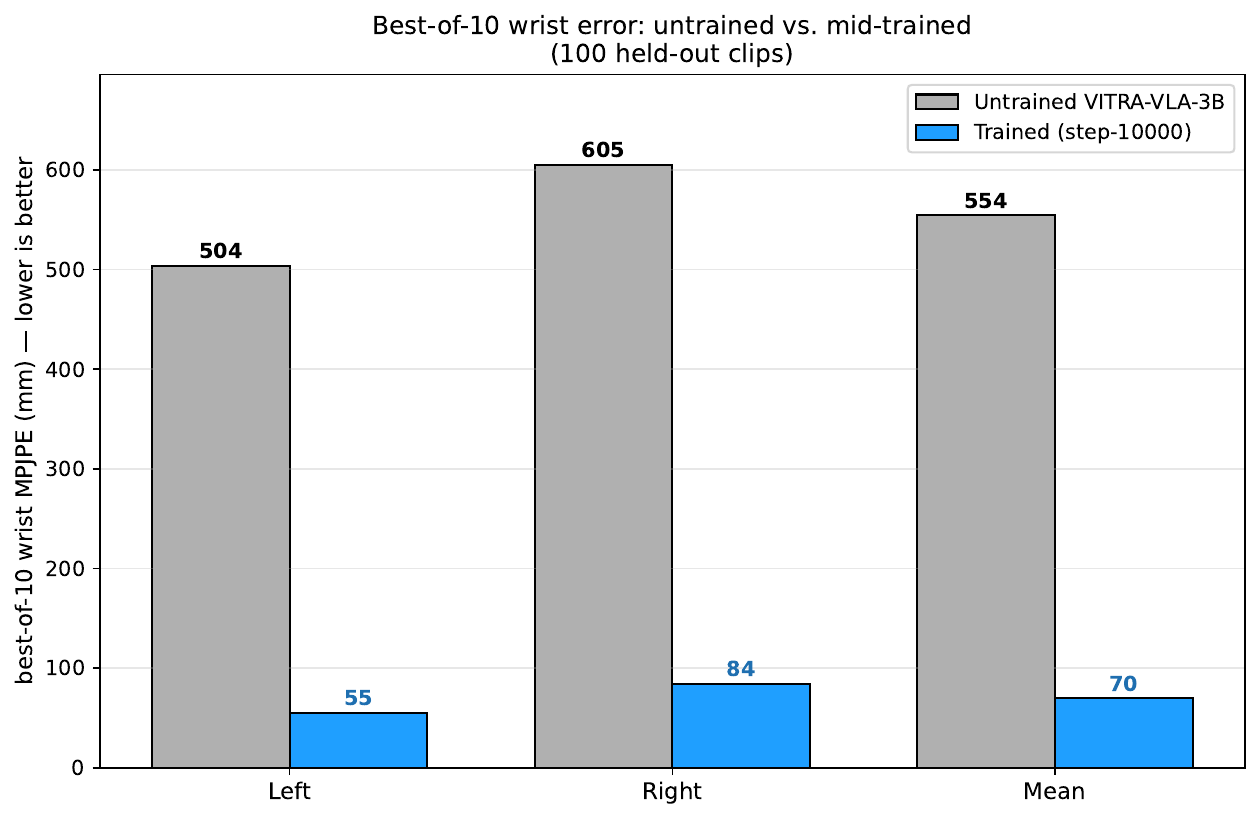}
        \caption{Best-of-10 wrist error on held-out clips: the untrained VITRA-VLA-3B is mis-scaled on our data (554\,mm); mid-training calibrates it to 70\,mm.}
        \label{fig:qualitative}
    \end{subfigure}
    \caption{Downstream VLA mid-training on MobileEgo Anywhere using the VITRA framework~\cite{vitra}.}
    \label{fig:downstream_vla_training}
\end{figure*}

To test whether MobileEgo Anywhere data is a usable training signal for action models, we mid-train the VITRA vision--language--action framework~\cite{vitra} on our egocentric trajectories and evaluate on a held-out split of 27 sessions disjoint from training. We initialize from VITRA-VLA-3B, freeze the vision encoder, condition on hierarchical instructions, and supervise MANO hand pose prediction for 10{,}000 steps. Following VITRA, we predict directly in MANO hand-parameter space (per-hand wrist 6-DoF and finger joint angles), which corresponds to anthropomorphic robot hands and avoids learning a human-to-robot retargeting during pretraining.

Held-out action-prediction loss decreases monotonically from 0.265 to 0.214 ($-19\%$; Fig.~\ref{fig:downstream_vla_training}a). Because the metric is computed on sessions never seen during training, this drop reflects generalization rather than memorization, and it is concentrated in finger articulation ($-24\%$ per hand) over the wrist root ($-14\%$), indicating that the model learns the fine, dexterity-relevant hand structure our labels carry. The untrained VITRA-VLA-3B does not transfer to our data out of the box: its predictions are mis-scaled, giving a best-of-10 wrist error of 554\,mm, whereas mid-training calibrates it to our distribution (70\,mm; Fig.~\ref{fig:downstream_vla_training}b) and reduces held-out loss from 1.00 to 0.21. These results establish open-loop trainability rather than policy-level accuracy: we evaluate hand-action prediction on held-out human sessions and perform no closed-loop robot rollout.

\section{Limitations}

\textbf{Platform dependency:} The capture pipeline currently requires an iPhone Pro, as STERA relies on ARKit's visual-inertial odometry and LiDAR depth sensing. Android/ARCore offers lower VIO accuracy and lacks LiDAR depth; supporting depth-free fallback modes would meaningfully broaden contributor reach. Additionally, the ultrawide lens is inaccessible during active ARKit sessions, limiting field of view for wide-workspace activities.

\textbf{Thermal constraints on session length} Continuous recording beyond approximately two hours can trigger thermal throttling on iPhone Pro hardware; a heat sink attachment mitigates this but adds deployment friction in warm environments.

\textbf{Preliminary downstream validation:} Our training-signal experiment is open-loop (we evaluate held-out hand-action prediction and perform no closed-loop robot rollout), so it demonstrates trainability rather than policy-level task success, which we leave to future work.
%Though iphone is a great capture device and it democratizes data access, there are a couple of limitations - challenges in using ultrawide images and heating issues during prolonged operation. Ultrawide images cannot be accessed during ARkit session and ultrawide image is often very useful for egocentric data collection due to its wider context. The device does heat up, when operated beyond 2+ hours, but this issue can be mitigated by placing a heat sink.

\section{Conclusion}

We have presented MobileEgo Anywhere, an accessible and commoditized framework for large-scale, long-horizon egocentric data collection. By releasing a free mobile application and open-sourcing the STERA processing pipeline, we enable researchers and contributors worldwide to generate VLA-ready datasets using standard consumer hardware. The 200-hour dataset features continuous episodes up to 108 minutes, 6\,DoF ARKit poses, LiDAR depth, MANO 3D hand trajectories in a consistent world frame, and three-level hierarchical language annotations. This lowers the barrier to creating VLA-ready egocentric datasets and supports work toward more generalizable robot policies.

\clearpage

\twocolumn[
  \begin{center}
    \centering
    \section*{Appendix} 
    \vspace{1em} % Optional: adjusts spacing below the centered title
  \end{center}
]

\begin{figure*}[!t] 
    \centering
    \includegraphics[width=0.95\textwidth]{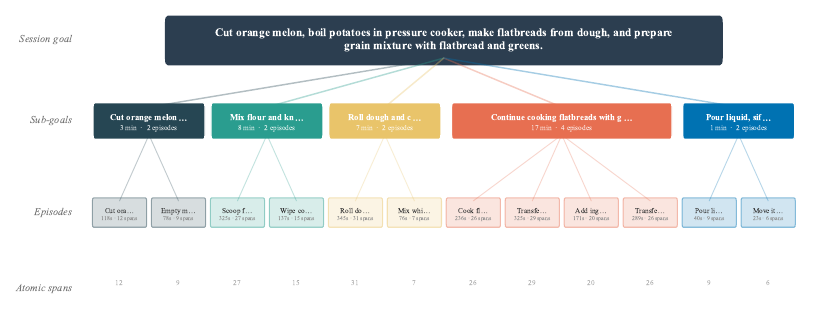}
    \caption{Hierarchical decomposition of a 36-minute cooking session (217 atomic spans). A single session goal decomposes into five sub-goals, each containing two to four episodes. Sub-goal durations range from 1 to 17 minutes; episode durations from 23\,s to 345\,s.}
    \label{fig:tree_example}
\end{figure*}

\section{3D Hand Pose Consistency}

As ground-truth MANO poses are unavailable at our scale, we assess quality across 98 sessions (1.19 M frames, 25.2 hours) via three ground-truth-free metrics: bone length constancy, joint angle plausibility, and wrist dynamics. Hand detection succeeds on 86.2\% of frames (mean WiLoR confidence 0.73); 247 frames with zero LiDAR depth returns (0.02\%) are discarded via a $z > 0.01$\,m threshold before computing metrics.

\begin{figure*}[!t] 
    \centering
    \begin{subfigure}[t]{0.48\textwidth}
        \centering
        \includegraphics[width=\textwidth]{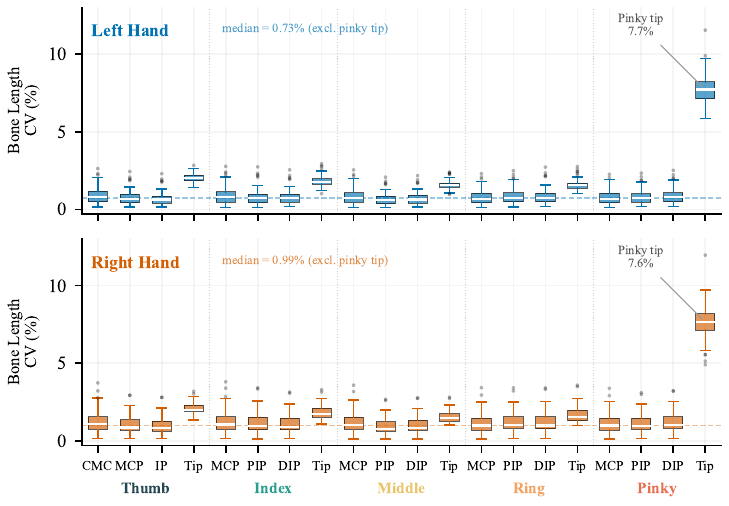}
        \caption{Per-bone Co-efficient of Variation (CV) of bone length over 98 sessions. The pinky distal bone shows elevated CV due to its short physical length ($\sim$2\,cm); excluding it, median CV falls below 1.}
        \label{fig:bone_cv}
    \end{subfigure}
    \hfill
    \begin{subfigure}[t]{0.48\textwidth}
        \centering
        \includegraphics[width=\textwidth]{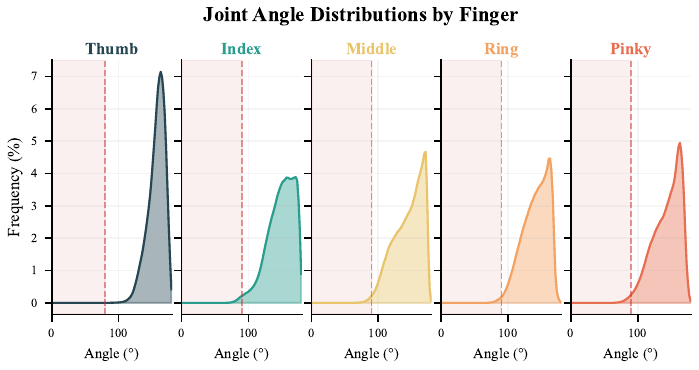}
        \caption{Joint flexion angle distributions pooled over 98 sessions. Shaded regions indicate biomechanical limits; $>$99.99\% of angles fall within bounds.}
        \label{fig:joint_angles}
    \end{subfigure}
    \caption{Kinematic evaluation results.}
    \label{fig:kinematic_evaluation}
\end{figure*}

\textbf{Bone length constancy.} Median coefficient of variation (CV) of the 20 MANO bone lengths is 1.27\% (left hand) and 1.43\% (right), indicating stability to within roughly 1\,mm on a typical 7--8\,cm bone (Fig.~\ref{fig:bone_cv}). The pinky distal phalanx shows elevated CV ($\sim$7.5\%) due to its short physical length ($\sim$2\,cm), which amplifies relative error from a fixed absolute noise floor; excluding it, pooled median CV drops below 1\% for both hands.

\textbf{Joint angle plausibility.} Over 99.99\% of the 15 flexion angles (MCP, PIP, and DIP for each finger) fall within published biomechanical limits~\cite{cobos2008hand} across all sessions (Fig.~\ref{fig:joint_angles}), with unimodal distributions consistent with the variety of grasp types present in the dataset.

\begin{figure*}[t] % Changed from [!t] to [t] for proper two-column top placement
    \centering
    % Left Figure
    \begin{subfigure}[t]{0.48\textwidth}
        \centering
        \includegraphics[width=\textwidth]{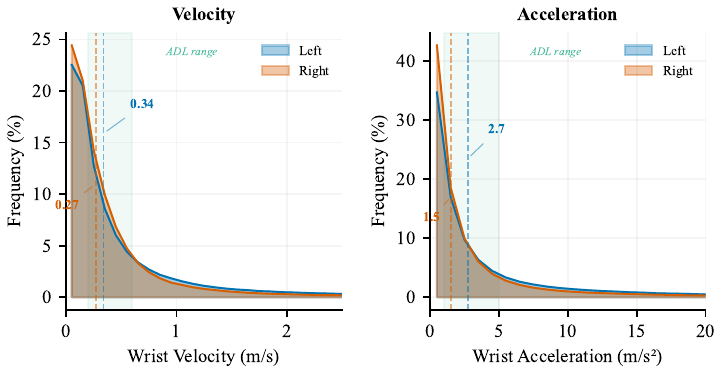}
        \caption{Wrist velocity and acceleration distributions for left and right hands, pooled over 98 sessions; median velocity is 0.34~m/s (left) and 0.27~m/s (right). These session medians fall below the $\sim$0.6--0.9\,m/s peak hand velocities reported for everyday reaching tasks~\cite{altmurphy2011drinking,nakatake2023drinking}.}
        \label{fig:wrist_dynamics}
    \end{subfigure}
    \hfill % Adds horizontal space to push the figures to the edges
    % Right Figure
    \begin{subfigure}[t]{0.48\textwidth}
        \centering
        \includegraphics[width=\textwidth]{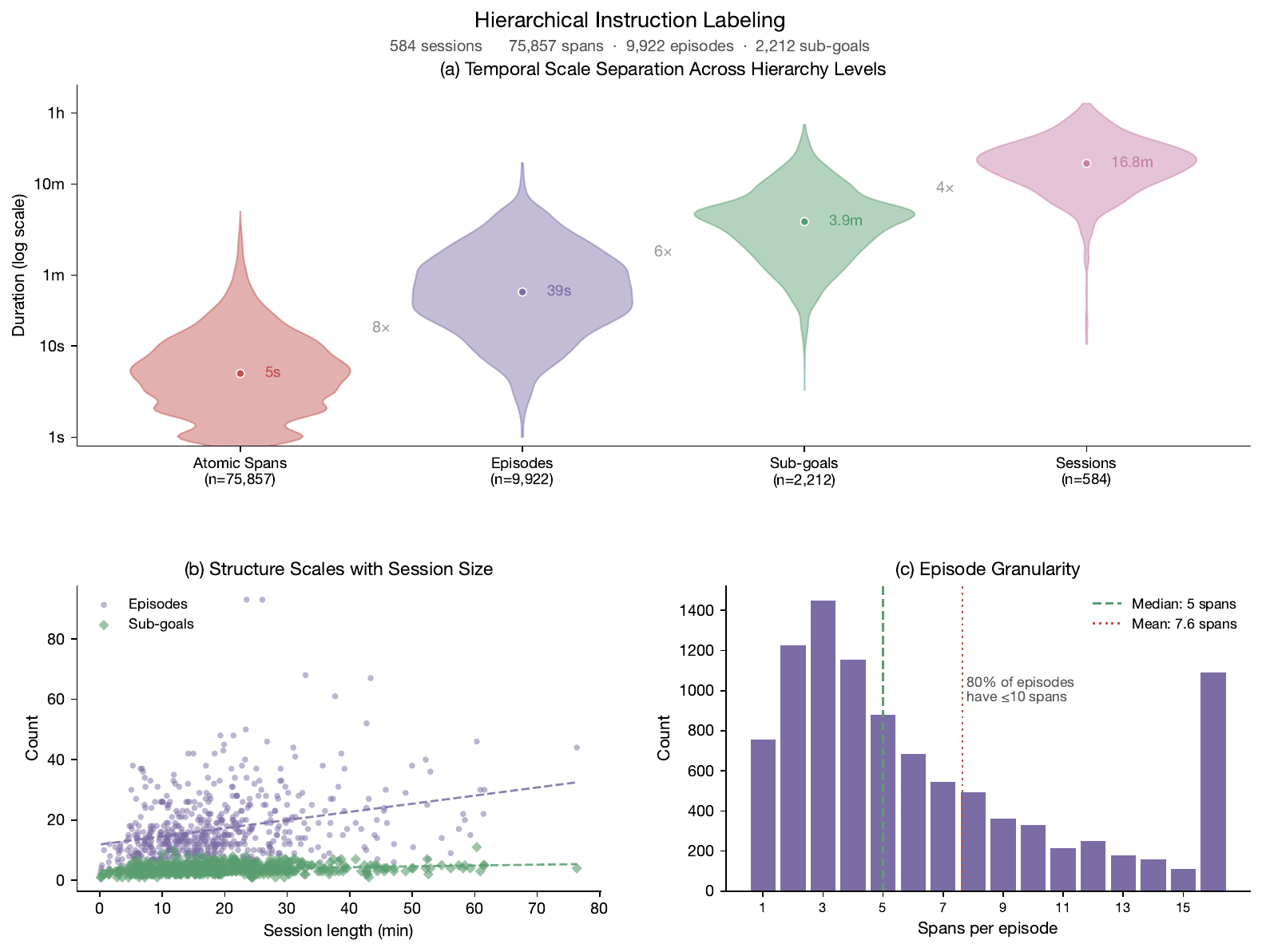}
        \caption{Hierarchical instruction analysis across 584 sessions (75,857 atomic spans). (1)~Temporal scale separation with consistent 4--8$\times$ gaps between levels. (2)~Linear scaling of episode and sub-goal counts with session length. (3)~80\% of episodes contain $\leq$10 atomic spans (median 5, mean 7.6).}
        \label{fig:hierarchical}
    \end{subfigure}
    
    \caption{Overview of experimental data: Left shows wrist dynamics distributions, while right details the hierarchical instruction analysis.}
    \label{fig:combined_metrics}
\end{figure*}

\textbf{Wrist dynamics.} Median wrist velocity is 0.34\,m/s (left) and 0.27\,m/s (right), and median acceleration is 2.7 and 1.5\,m/s$^2$ (Fig.~\ref{fig:wrist_dynamics}). The velocity medians sit below the $\sim$0.62\,m/s peak hand velocity reported for healthy adults during the standardized drinking task~\cite{altmurphy2011drinking,nakatake2023drinking}, as expected for continuous recordings that interleave fast reach-and-transport phases with slower in-hand manipulation; the smooth, unimodal distributions show no teleportation-scale discontinuities.

\section{Hierarchical Instruction Quality}

We ran the hierarchical decomposition across all 584 sessions using DeepSeek V4 Flash with high reasoning, producing 75,857 atomic spans grouped into 9,922 episodes and 2,212 sub-goals. Three structural invariants are enforced during generation: every atomic span maps to exactly one episode, episode and sub-goal boundaries use exact input timestamps, and each decomposition covers the full session with no gaps.

\begin{figure*}[p] 
    \centering
    
    % --- Row 1 ---
    \begin{subfigure}[t]{0.27\textwidth}
        \centering
        \includegraphics[width=\textwidth]{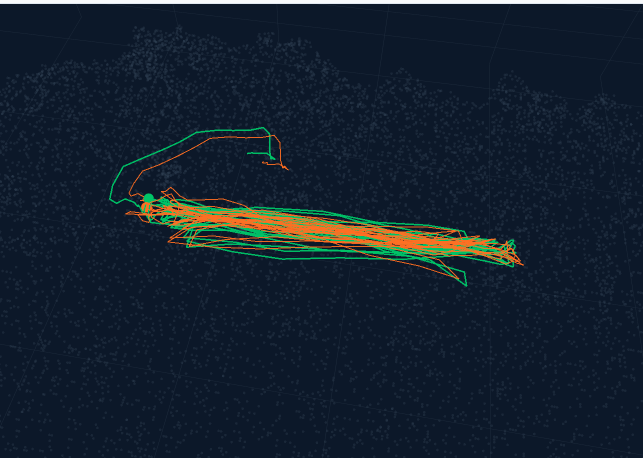}
        \caption{Session 1 - Repeated traversals along a fixed straight-line path between static reference markers at three distinct locomotion speeds.}
        \label{fig:s1}
    \end{subfigure}
    \hfill
    \begin{subfigure}[t]{0.32\textwidth}
        \centering
        \includegraphics[width=\textwidth]{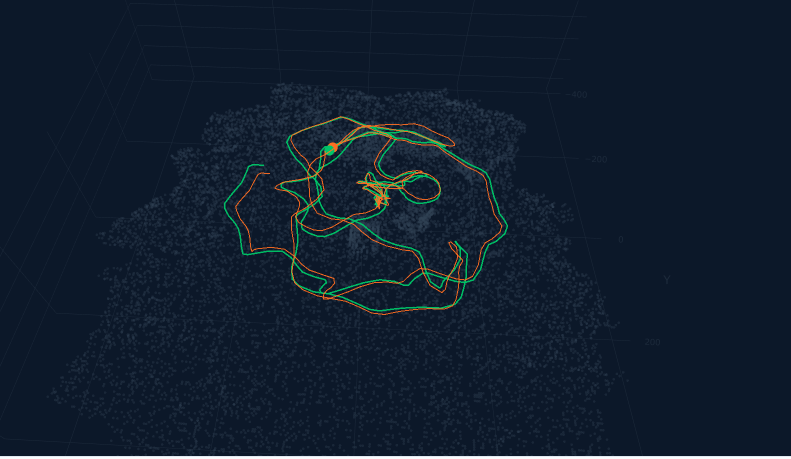}
        \caption{Session 2 - Slow exploration with gentle head rotations, dynamic distractors, seated interaction, and partial occlusions.}
        \label{fig:s2}
    \end{subfigure}
    \hfill
    \begin{subfigure}[t]{0.32\textwidth}
        \centering
        \includegraphics[width=\textwidth]{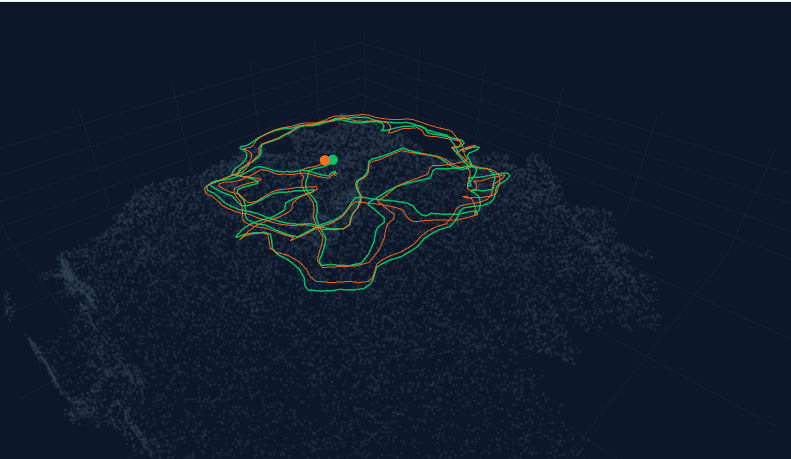}
        \caption{Session 3 - Slow exploration with head rotations and dynamic distractors, without seated interaction, isolating low-speed navigation.}
        \label{fig:s3}
    \end{subfigure}

    \vspace{0.3cm}

    % --- Row 2 ---
    \begin{subfigure}[t]{0.32\textwidth}
        \centering
        \includegraphics[width=\textwidth]{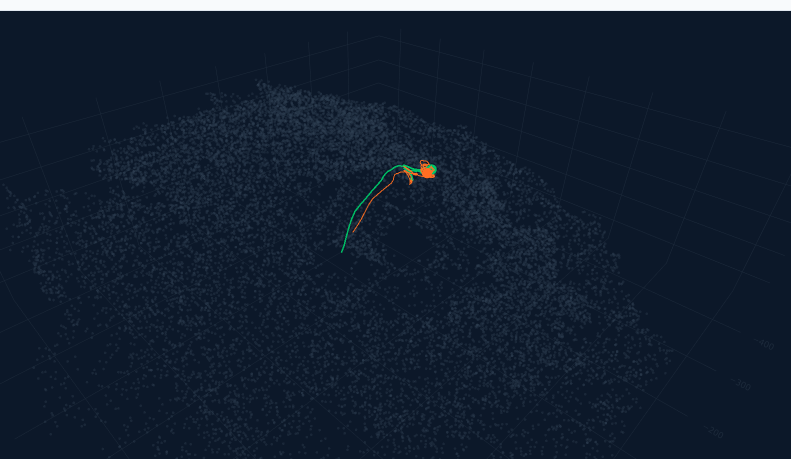}
        \caption{Session 4 - Slow continuous rotation while seated on a swivel chair, producing sustained pure-yaw motion.}
        \label{fig:s4}
    \end{subfigure}
    \hfill
    \begin{subfigure}[t]{0.32\textwidth}
        \centering
        \includegraphics[width=\textwidth]{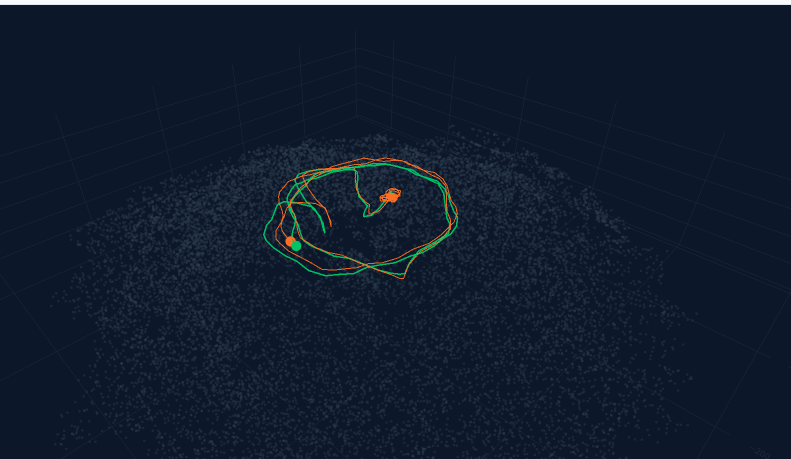}
        \caption{Session 5 - Aggressive rotation while seated on a swivel chair, including rapid full revolutions and high angular velocity.}
        \label{fig:s5}
    \end{subfigure}
    \hfill
    \begin{subfigure}[t]{0.32\textwidth}
        \centering
        \includegraphics[width=\textwidth]{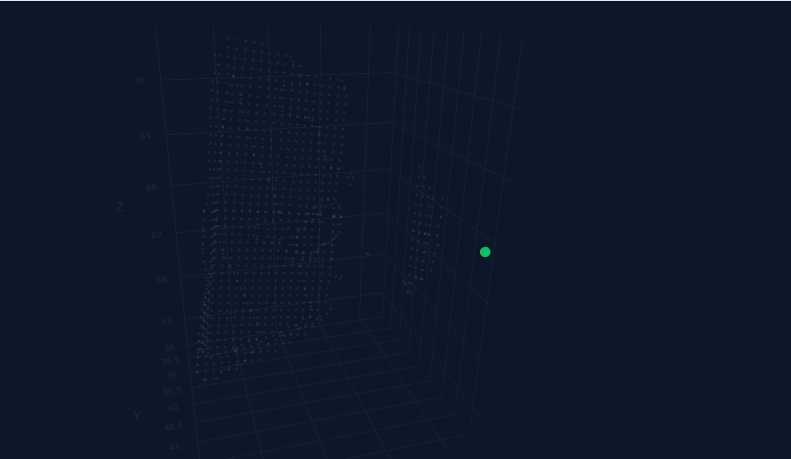}
        \caption{Session 6 - Fully stationary recording with no intentional motion, used to quantify positional and rotational drift at rest.}
        \label{fig:s6}
    \end{subfigure}

    \vspace{0.3cm}

    % --- Row 3 ---
    \begin{subfigure}[t]{0.32\textwidth}
        \centering
        \includegraphics[width=\textwidth]{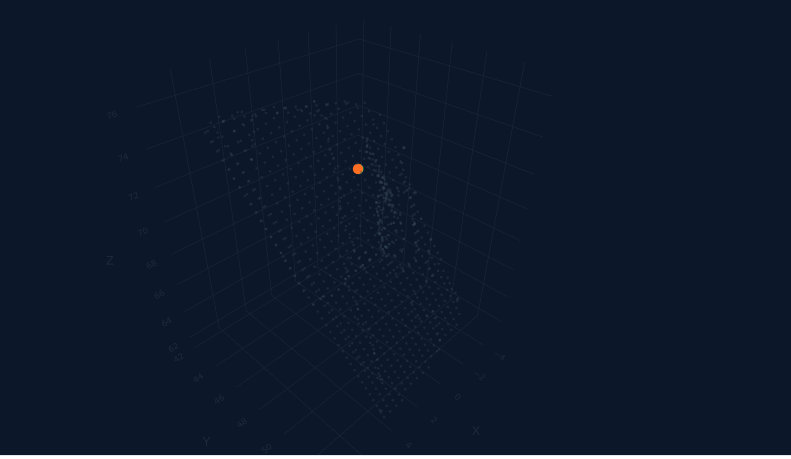}
        \caption{Session 7 - A second fully stationary recording with no intentional motion, providing a repeated measurement of static drift.}
        \label{fig:s7}
    \end{subfigure}
    \hfill
    \begin{subfigure}[t]{0.32\textwidth}
        \centering
        \includegraphics[width=\textwidth]{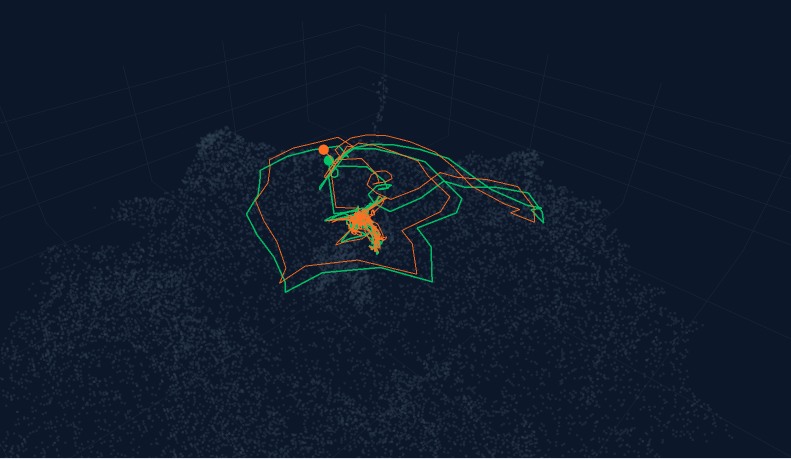}
        \caption{Session 8 - Egocentric activity sequence involving seated object interaction (opening bags, handling cameras, folding clothes).}
        \label{fig:s8}
    \end{subfigure}
    \hfill
    \begin{subfigure}[t]{0.32\textwidth}
        \centering
        \includegraphics[width=\textwidth]{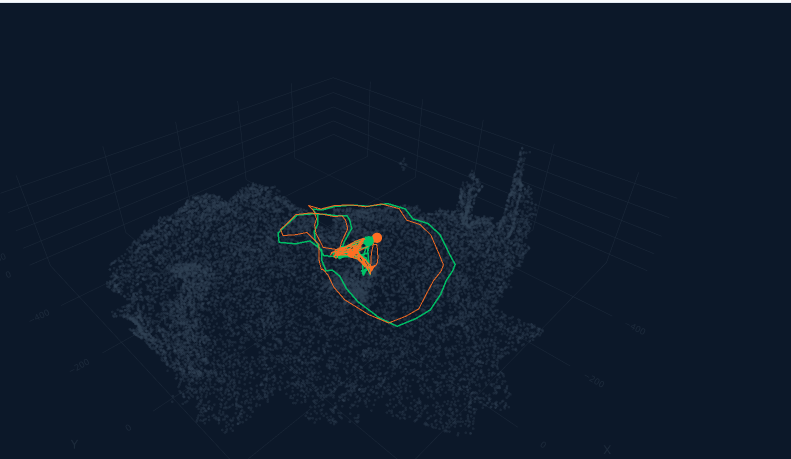}
        \caption{Session 9 - Activity interactions performed at higher speed and with abrupt movements, increasing motion blur.}
        \label{fig:s9}
    \end{subfigure}

    \vspace{0.3cm}

    % --- Row 4 ---
    \begin{subfigure}[t]{0.32\textwidth}
        \centering
        \includegraphics[width=\textwidth]{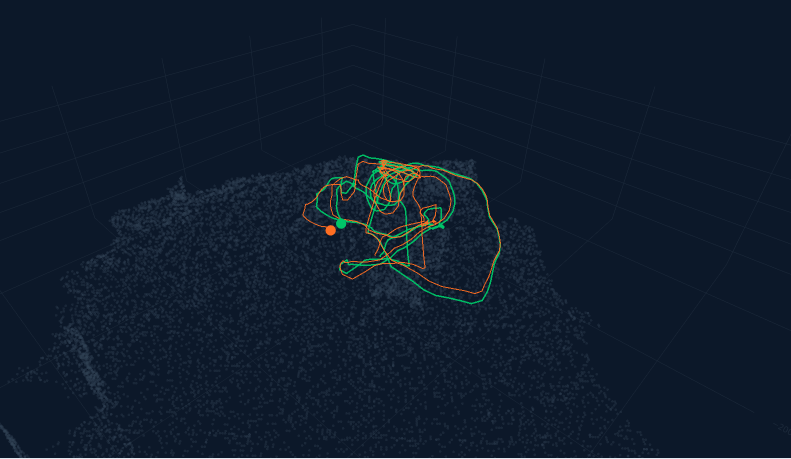}
        \caption{Session 10 - An activity in which the person climbed a stool, adding height variation to the trajectory.}
        \label{fig:s10}
    \end{subfigure}
    \hfill
    \begin{subfigure}[t]{0.20\textwidth} 
        \centering
        \includegraphics[width=\textwidth]{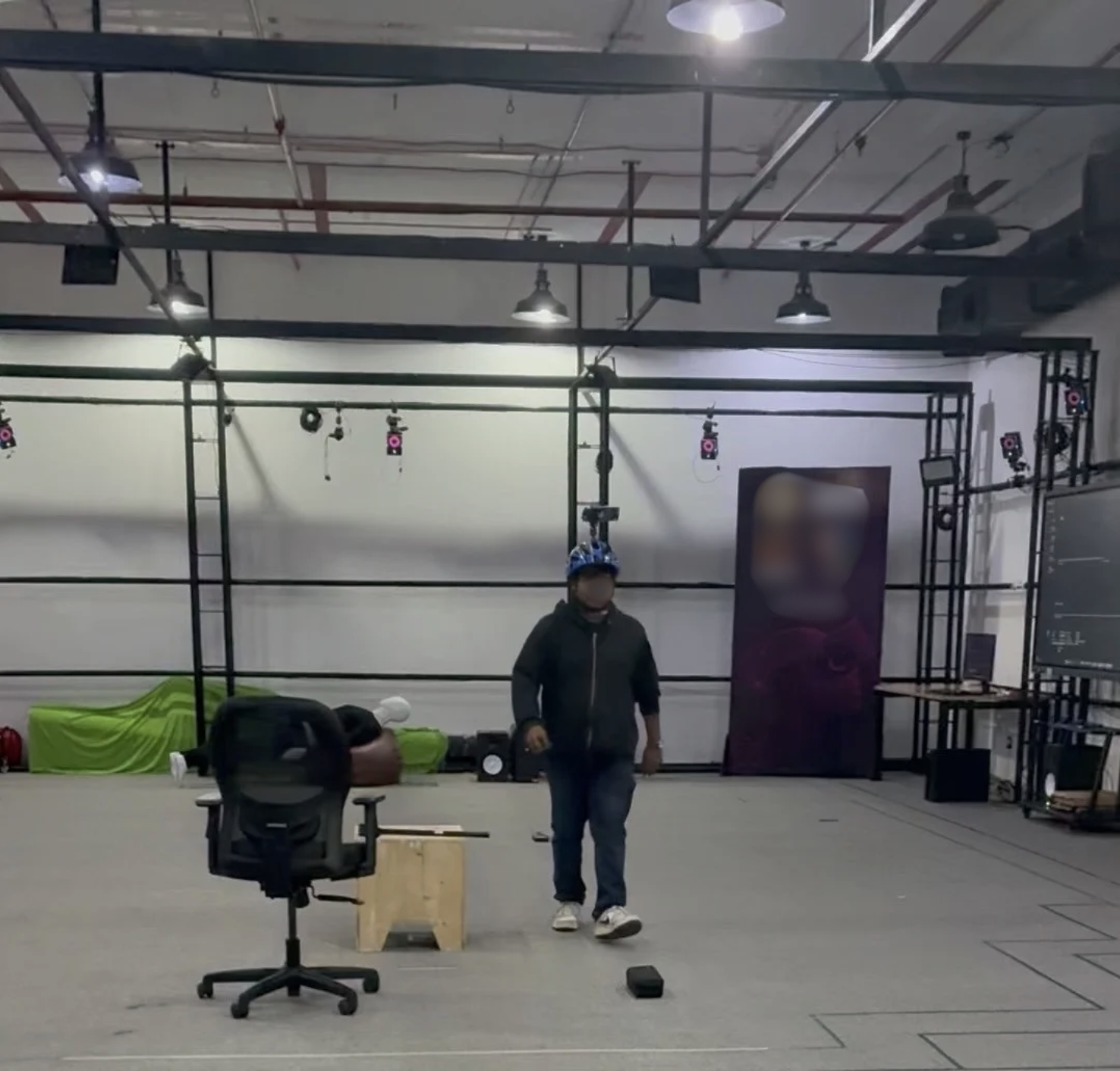}
        \caption{Mocap setup layout.}
        \label{fig:mocap_setup}
    \end{subfigure}
    \hfill
    \begin{subfigure}[t]{0.32\textwidth}
        \centering
        \includegraphics[width=\textwidth]{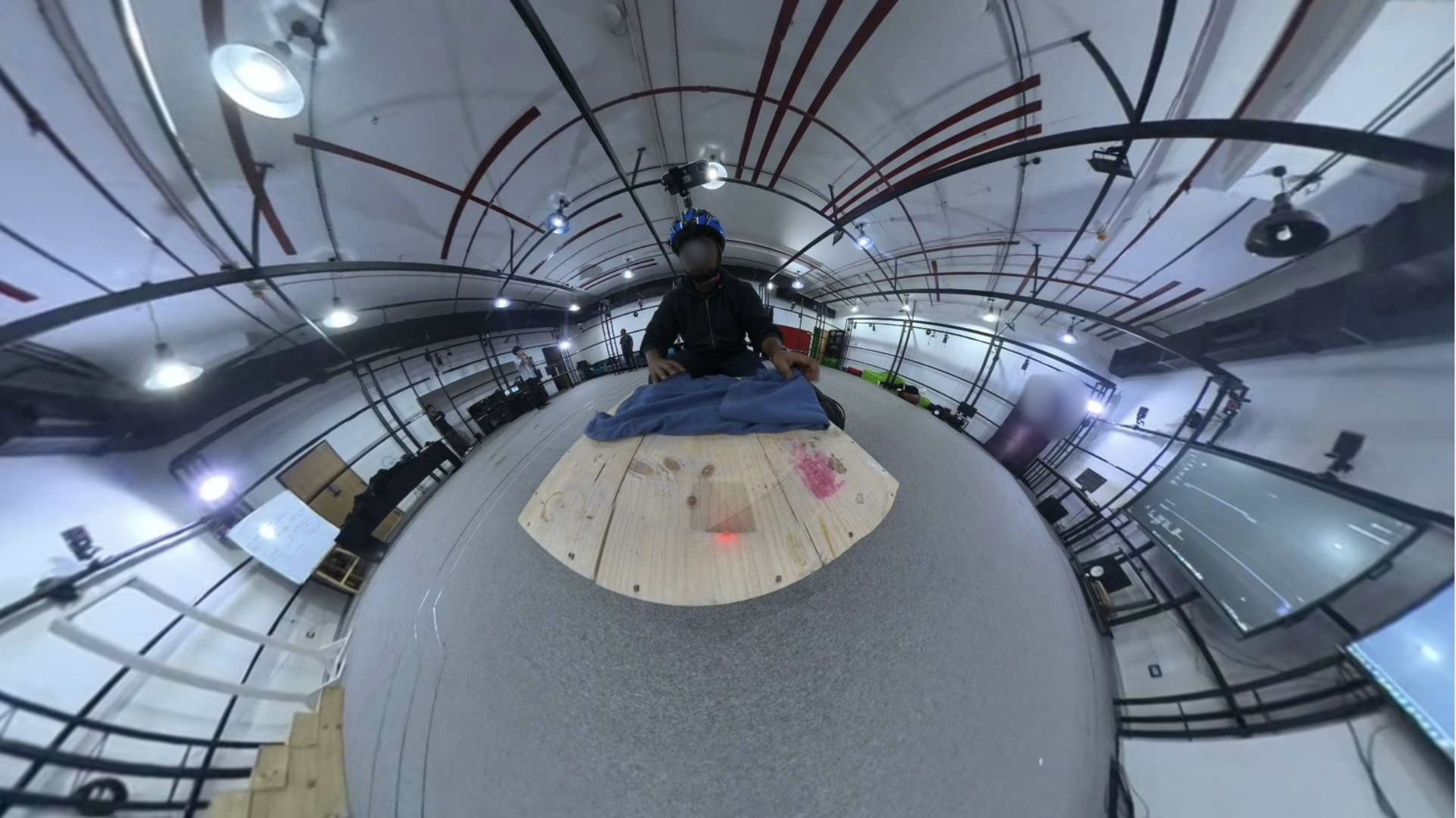}
        \caption{A contributor performing an egocentric activity while tracked by the motion-capture system.}
        \label{fig:user_activity}
    \end{subfigure}
    
    \vspace{0.2cm}
    \caption{Unified overview of Mocap trajectory vs ARKit trajectory across sessions 1-10 (green: mocap, orange: ARKit) alongside the corresponding hardware capture environment and setup layout.}
    \label{fig:all_sessions_combined}
\end{figure*}

Fig.~\ref{fig:all_sessions_combined} shows that each level of the hierarchy occupies a distinct temporal band, with a natural 4--8$\times$ scale separation between adjacent levels: 5\,s atomic spans, 39\,s episodes, 3.9\,min sub-goals, and 16.8\,min sessions. This structure emerges from the data rather than being imposed by the prompt. Episode and sub-goal counts scale roughly linearly with session length (Fig.~\ref{fig:all_sessions_combined}), confirming adaptation to session complexity. Most episodes are compact: 80\% contain $\leq$10 atomic spans (Fig.~\ref{fig:all_sessions_combined}), with a median of 5 spans per episode. The total LLM cost for hierarchical structuring was only a few dollars, negligible relative to the data-collection effort. A representative decomposition of a 36-minute session is shown in Fig.~\ref{fig:tree_example}.

\section{Mocap trajectories and results}

The trajectories captured in motion capture settings for validating arkit accuracies, discussed in ~\ref{mocap_arkit} is shown in Figure ~\ref{fig:all_sessions_combined}
% Text insertion or manually authored analysis can be added here if needed.

\section{Ethics and Privacy}

All contributors signed informed consent covering capture, processing, and public release. Contributors were instructed to avoid recording non-consenting individuals; any faces accidentally captured were blurred in post-processing using \cite{egoblur}

\begin{figure*}[!t]
    \centering
    \includegraphics[width=0.95\textwidth]{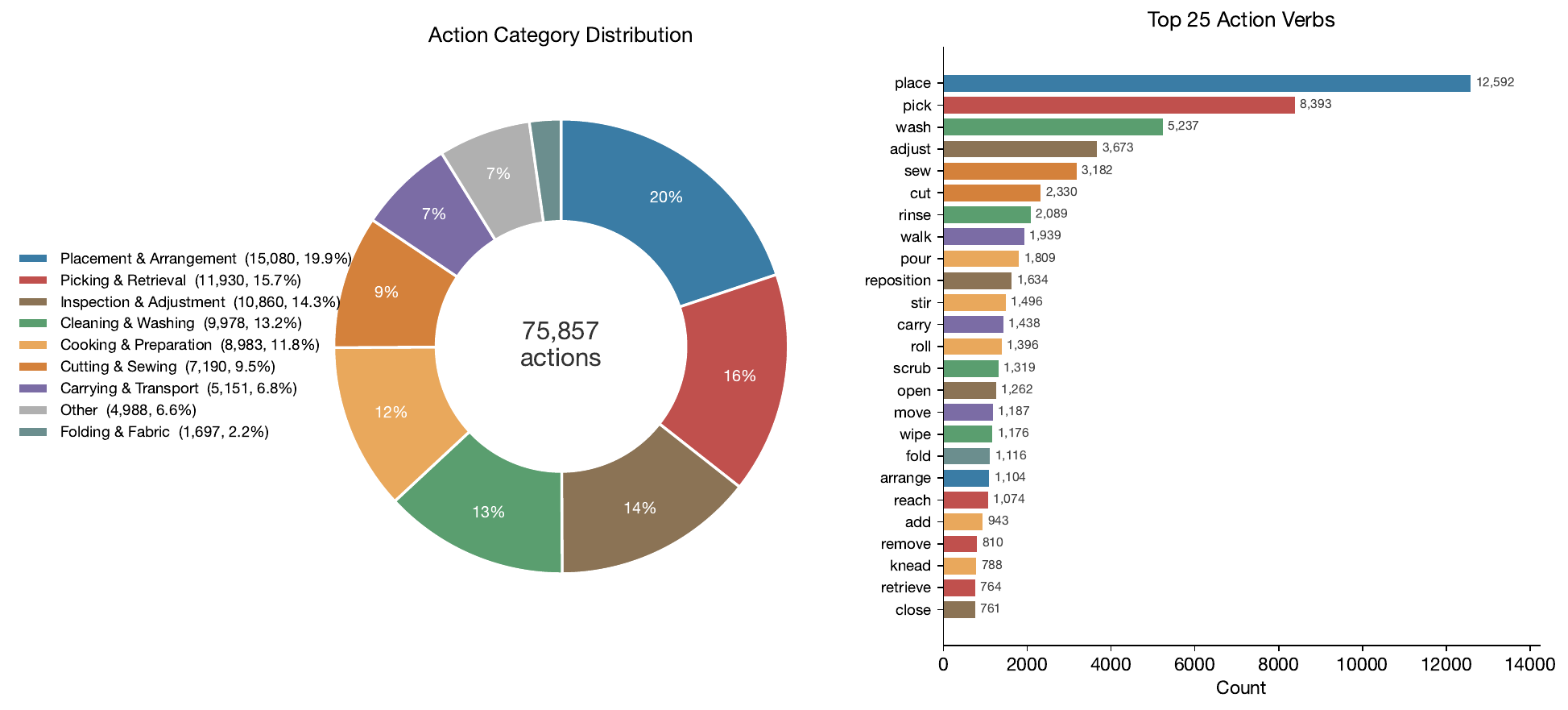}
    \caption{Task diversity across 584 sessions and 20 contributors. The 75{,}857 atomic action labels span a long-tailed vocabulary over household manipulation domains (cooking, cleaning, sewing, organizing), grouped into nine high-level categories.}
    \label{fig:task_diversity}
\end{figure*}

\end{document}